\documentclass[journal]{IEEEtran}
\usepackage{times}
\usepackage{epsfig}
\usepackage{graphicx}
\usepackage{amsmath}
\usepackage{amssymb}

\usepackage{cite}
\usepackage{enumitem}

\usepackage[ruled]{algorithm2e}

\usepackage{multirow}
\usepackage{makecell}
\usepackage{booktabs}
\usepackage{color}
\usepackage{threeparttable}

\newcommand{\etal}{\textit{et al.}}
\newcommand{\ie}{\textit{i.e.}}
\newcommand{\eg}{\textit{e.g.}}
\definecolor{purple}{rgb}{0.54, 0.17, 0.89}
\def\red{\textcolor{red}}

\def\green{\textcolor{green}}

\ifCLASSINFOpdf
\else
\fi
\begin{document}
\title{MambaTrack3D: A State Space Model Framework for LiDAR-Based Object Tracking under High Temporal Variation}

\author{
        Shengjing Tian,
        Yinan Han,
        Xiantong Zhao*,
        Xuehu Liu,
        and~Qi Lang
\thanks{Manuscript received --; revised --}
\thanks{Shengjing~Tian is with the School of Economics and Management, China University of Mining and Technology, China. tye.dut@gmail.com }
\thanks{Yinan Han is with the DUT-BSU Joint Institute, Dalian University of Technology, China. Spolico\_hyn@outlook.com}
\thanks{Xiantong Zhao are with the School of Mathematical Sciences, Dalian University of Technology, China. Corresponding author: xtongz.dut@gmail.com}
\thanks{Qi Lang are with the School of Information Science and Technology, Northeast Normal University, China. langqi@nenu.edu.cn}
\thanks{Xuehu Liu are with the School of Computer Science and Artificial Intelligence, Wuhan University of Technology, China. liuxuehu@whut.edu.cn}
}

% Shengjing Tian: Conceptualization, methodology, validation
% formal analysis, visualization, writing;
% Jun Liu: Conceptualization, formal analysis, review and editing;
% Xiuping Liu: Conceptualization, review and editing, supervision.


\markboth{Journal of \LaTeX\ Class Files,~Vol.~14, No.~8, July~2021}%
{Shell \MakeLowercase{\textit{et al.}}: Bare Demo of IEEEtran.cls for IEEE Journals}



\maketitle

\begin{abstract}
Dynamic outdoor environments with high temporal variation (HTV) pose significant challenges for 3D single object tracking in LiDAR point clouds. Existing memory-based trackers often suffer from quadratic computational complexity, temporal redundancy, and insufficient exploitation of geometric priors. To address these issues, we propose MambaTrack3D, a novel HTV-oriented tracking framework built upon the state space model Mamba. Specifically, we design a Mamba-based Inter-frame Propagation (MIP) module that replaces conventional single-frame feature extraction with efficient inter-frame propagation, achieving near-linear complexity while explicitly modeling spatial relations across historical frames. Furthermore, a Grouped Feature Enhancement Module (GFEM) is introduced to separate foreground and background semantics at the channel level, thereby mitigating temporal redundancy in the memory bank. Extensive experiments on KITTI-HTV and nuScenes-HTV benchmarks demonstrate that MambaTrack3D consistently outperforms both HTV-oriented and normal-scenario trackers, achieving improvements of up to +6.5 success / +9.5 precision over HVTrack under moderate temporal gaps. On the standard KITTI dataset, MambaTrack3D remains highly competitive with state-of-the-art normal-scenario trackers, confirming its strong generalization ability. Overall, MambaTrack3D achieves a superior accuracy–efficiency trade-off, delivering robust performance across both specialized HTV and conventional tracking scenarios.
\end{abstract}


% \fi
\IEEEpeerreviewmaketitle

\section{Introduction}
The domain of visual information processing has been transitioning from two-dimensional (2D) camera pixels to three-dimensional (3D) geometric point clouds \cite{ipm_pvstrans, ipm_voxelGNN}, wherein Light Detection and Ranging (LiDAR) has been serving as a predominant sensor. Accordingly, visual object tracking (VOT) from the LiDAR point clouds also has been becoming indispensable to perceive the surrounding environments in the rapidly evolving fields of robotic vision \cite{Comport2004RobustMT, Machida2012HumanMT}, autonomous driving \cite{Argoverse2019, KITTI2013}, and augmented reality \cite{tracking_AR}. The primary goal of the VOT is to employ a bounding box to estimate the pose and centroid of the object of interest within frame streams, facilitating systems in the real-time localization capacity within the 3D physical space. Recent research has increasingly focused on point clouds for 3D object detection and tracking \cite{ipm_voxelGNN, nuscenes}. However, the majority of existing literature emphasizes accuracy in normative settings \cite{P2B} \cite{stnet} \cite{MBPTrack}, often overlooking special scenarios where long intervals exist between two adjacent perceptions, called  High Temporal Variation (HTV).

\begin{figure}
    \centering
    \includegraphics[width=1\linewidth]{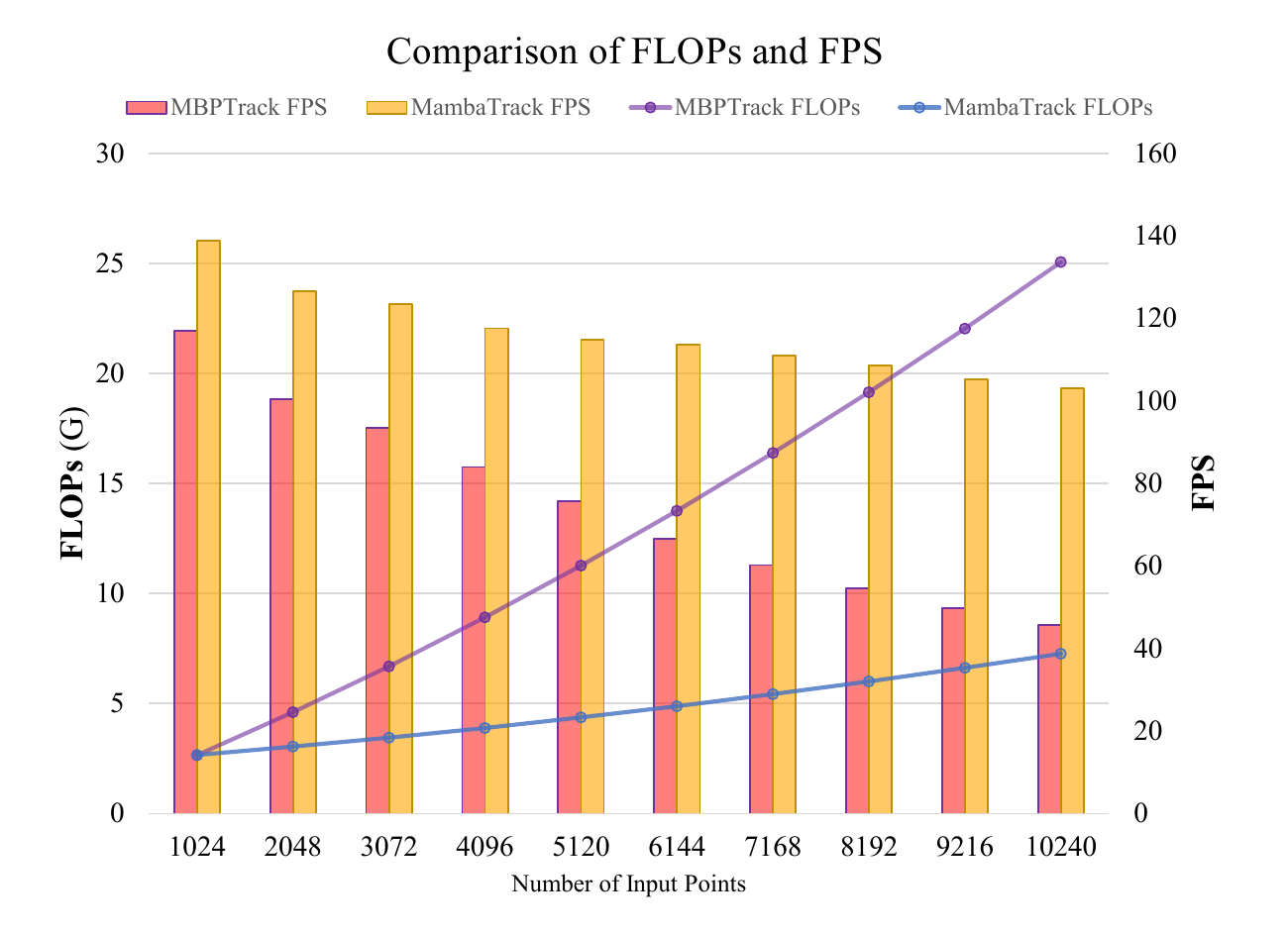}
    \caption{Performance comparisons between transformer-based method and our method. The X-axis is the number of input points. The FLOPs on the left Y-axis reflects the computation complexity, while the FPS on the right Y-axis represents the running speed of different methods.  }
    \label{fig:flops}
\end{figure}
HTV tracking plays a pivotal role in real-world applications, as it can markedly decrease computational demands across a wide range of tasks, such as detection and segmentation, through the adoption of skipped-tracking strategies \cite{FlexPatch, EdgeAR}. Moreover, HTV tracking is essential for deployment on edge devices, such as Unmanned Aerial Vehicles, which are often constrained by limited point cloud resolution, computational capability, and power supply \cite{EdgeAR, power_save}. To this end, the approaches for HTV tracking must be sufficiently efficient to tackle the following challenges: High Computational Efficiency (HCE), Large Appearance Variations (LAV), Distractions from Similar Objects (DSO), and Background Noise (BN). The HCE demands that the tracking algorithm operate at high frame rates to meet real-time requirements. The LAV affects the point density and spatial distribution of LiDAR data, typically arising from occlusions, rapid object motion, and environmental factors such as weather changes, which may cause objects to be partially or fully obscured. The DSO is also inevitable because we necessitate an expansion of the 3D search area to accommodate substantial movements. This expansion increases the likelihood of interference from visually similar objects. Naturally, the enlargement of the search area inherently reduces the proportion of target-relevant information and amplifies the presence of background noise (BN) within the scene.  

Current research in LiDAR VOT primarily leverages the Siamese paradigm \cite{P2B,Siam3D2019} to improve accuracy under normative settings. This framework takes a target template from the previous frame and a search area from the current frame as inputs, encoding them into a latent feature space to capture both global and local semantic information. It then renders the target surface points discriminative by integrating template information into the search area, ultimately followed by a localization network to estimate the target's state in the current frame. Although methods like P2B \cite{P2B}, STNet \cite{stnet}, and CXTrack \cite{cxtrack} have made significant strides, they often struggle in high temporal variation scenarios. This limitation arises because these approaches only pass the target cue from the latest frame to the current one, without effectively leveraging historical frame information. To address this, several memory-based trackers \cite{TAT}\cite{SCTrack}\cite{SeqTrack3D}\cite{MBPTrack} have been proposed to perform tracking in point clouds. Among them, the most outstanding and representative tracker, MBPTrack \cite{MBPTrack}, utilizes the Transformer \cite{attention} to process the rich temporal context across historical frames. 
However, under HTV conditions, these methods still face challenges in efficiently managing and utilizing sequential data. In light of this, HVTrack \cite{hvtrack} specifically targets HTV scenarios by designing dedicated cross-attention and self-attention modules within a memory bank to deal with large appearance variations, similar object distractions, and background noise.  
Despite the advancements achieved by memory-based approaches, they exhibit limitations in terms of High Computational Efficiency (HCE) and temporal redundancy. Specifically, attention modules are detrimental to the HCE, as they incur quadratic computational complexity, thereby hampering runtime performance as the memory bank grows. Moreover, the aggregation of multiple historical frames tends to yield dense point clouds within the memory bank, introducing temporal redundancy and weakening the correlation between the template and the search region. Furthermore, transformer-based matching strategies in memory banks rely merely on the cosine similarity of point cloud features across various frames, frequently overlooking geometric priors of objects---such as spatial relationships among historical frames---that are crucial for precise tracking.

%为了解决上述问题，本文针对高时序变化场景提出了一种基于状态空间模型Mamba的点云跟踪框架。具体地，针对高时态跟踪的高计算效率要求，我们创新性地设计了一种时空感知的帧间特征传播模块。该模块具有两个优点：1）不同于以往方法中使用的无参考的特征提取，比如直接使用pointnet++或DGCNN，该模块改用历史帧间特征传播的方式来代替特征提取，并在传播过程中考虑当前帧点云与历史帧点云的空间位置关系；2）不同于Transformer中的二次计算复杂度，该模块基于Linear-Time Sequence Model，这能够显著提升方法的计算效率，可达110+FPS，是以往HTV跟踪方法的4倍。针对记忆池中存在的时序信息冗余问题，我们给出了一种前景与背景分组协同的语义增强模块。该模块背后的动机为同一特征通道只负责一种类别的点，要么前景，要么背景\cite{}。因此我们将特征分为两组，并通过不同的线性变换来分别负责前景点与背景点。基于此，当前帧点云在与记忆池中的稠密点云进行correlation操作时缓解了来自不同帧的不同种类点的干扰。我们将所提出的方法应用在KITTI的高时序变化数据上，跟踪结果获得了显著的提升。
To address the aforementioned challenges, this paper proposes a point cloud tracking framework for high temporal variation (HTV) scenarios, built upon the state space model Mamba \cite{mamba}. Specifically, to meet the demand for high computational efficiency in HTV tracking, we develop a novel spatiotemporal-aware inter-frame feature propagation module. This module offers two key advantages:  
1) Unlike previous methods that rely on reference-free feature extraction(\eg, directly applying PointNet++ or DGCNN), our module replaces feature extraction with inter-frame feature propagation, explicitly modeling the spatial relationships between the current and historical point clouds during the propagation process.  
2) In contrast to the quadratic computational complexity of Transformer-based architectures, our module is based on a linear-time sequence model, which can achieve over 110 FPS and is four times faster than existing HTV tracking methods.
To further mitigate the temporal redundancy in the memory bank, we propose a grouped feature enhancement module that exploits foreground–background collaboration. The resulting features can facilitate distinguishing the correlated points between the current frame and the memory bank, further alleviating temporal redundancy caused by aggregating historical frames. We evaluate the proposed method on the KITTI dataset under high temporal variation settings, and the tracking results demonstrate significant performance improvements.

To sum up, the contributions of this study are threefold:
\begin{itemize}
    \item A more efficient tracking framework is proposed for 3D HTV tracking, which utilizes the state space model to revolute the conventional single-frame feature extraction into multiple-frame feature propagation. 
    \item A grouped feature enhancement module is proposed to capture foreground and background semantics, which leverages two grouped channel experts to learn discriminative features and degrade temporal redundancy.
    \item A superior accuracy–speed trade-off is achieved, with the proposed method delivering considerable accuracy gains and a 2× runtime improvement in HTV scenarios over the current state-of-the-art, while maintaining competitive performance under normal settings. 
\end{itemize}

\section{Related Work} 
% Lidar点云的目标跟踪由于其广泛的应用场景已经受到研究者广泛关注。本节将依次对常规跟踪方法 、面向特定环境的跟踪方法进行回顾。最后，我们对最优传输理论在视觉领域的应用进行回顾
Visual object tracking from the LiDAR point clouds has attracted  much attention from researchers, attributed to its wide range of applicable scenarios. In this section, we systematically examine 3D tracking in a normal setting, followed by a discussion of 3D tracking in a specific setting. Lastly, we review the application of optimal transport theory within the domain of vision.

% 常规跟踪方法。针对室外场景LIDAR数据的感知，Giancola等人率先利用pointnet计算候选点云片段的特征，并根据外观余弦相似性来跟踪目标。此后，这类问题的发展结合了多项技术以提高跟踪的精确度。
% Hui \etal \cite{stnet}提出了Siamese Transformer网络，其改进了局部特征提取方式。 Shan \etal \cite{PTT}则利用 Transformer networks 来优化特征融合的方式，利用交叉注意力来融合目标信息。Park\etal \cite{graphTrack} 则借助图神经网络在模板特征和搜索区域特征之间进行信息传播。考虑到 bird's eye view在捕捉物体运动的能力，Luo \etal \cite{PTTR}将BEV作为特征表示的辅助分支来提升跟踪进度，Hui\etal \cite{V2B}则在定位模块中将原有的vote-based方法改为BEV为基础的方法，其上每个特征点作为一个锚点有利于候选包围盒的生成。

\subsection{3D Tracking in Normal Settings} 
For the understanding of LiDAR data in outdoor environments, Giancola \etal \cite{Siam3D2019} pioneered the use of pointnet \cite{PointNet2017} to compute the appearance features of candidate point cloud segments and track the target based on the Siamese framework. Subsequent advancements in this field have incorporated sophisticated methodologies aimed at enhancing tracking accuracy. Within this conceptual paradigm, P2B \cite{P2B} accomplishes an end-to-end architecture design, utilizing a deep Hough vote-based localization \cite{DeepHough2019} and feature fusion between the target and the search area. BAT \cite{BAT} augments the fusion by employing box clouds to derive a nine-dimensional representation for each raw point to the vertex of the bounding box. 
Researchers have subsequently focused on three key aspects: feature extraction, feature fusion, and localization mechanisms. Hui \etal \cite{stnet} introduced a Siamese Transformer network which improves the local feature extraction method. Conversely, Shan \etal \cite{PTT} employed Transformer networks to integrate target information in feature fusion through cross-attention mechanisms. Park \etal \cite{graphTrack} utilized the graph neural network to propagate information between template and search area features. Considering the bird's-eye view's (BEV) capability in capturing motion, Luo \etal \cite{PTTR} utilized the BEV as an ancillary branch for feature representation to enhance the tracking process. Hui \etal \cite{V2B} focused on the localization module and replaced the traditional vote-based method with a BEV-based response map, where each ``pixel'' acts as an anchor point to facilitate the generation of candidate enclosing boxes. Xu \etal \cite{cxtrack} and Guo \etal \cite{CMT} posited that the spatial relationship of the target with respect to other objects within the background can serve as referential clues to mitigate the influence of interfering objects.

In addition to the above appearance-based methods, motion-based approaches predict the approximate center of the target in the current frame utilizing historical frame data, which can serve as prior knowledge to enhance the precision of final bounding-box predictions. M2Tracker \cite{MMTrack} and P2P\cite{p2p} utilizes single-frame historical data in conjunction with a template to estimate the motion offset between frames. In contrast, DMT \cite{DMT} utilizes multi-frame historical data and integrates it with the target template, generating the target's bounding box by deep Hough voting \cite{DeepHough2019}.

While recent breakthroughs have made notable progress in the LiDAR VOT, existing methods often struggle in scenarios characterized by large appearance changes. This limitation stems from the propensity of these methods to solely transfer the target cue from the previous frame to the current frame, neglecting to harness the rich contextual information available in historical frames. To overcome this shortcoming,a new class of memory-based trackers has emerged \cite{TAT}\cite{SCTrack}\cite{SeqTrack3D}\cite{MBPTrack}, specifically designed to leverage temporal context to enhance tracking performance in point cloud. TAT \cite{TAT} employs historical templates to develop target-specific features through processes of sampling, enhancement, and aggregation operations. MBPTrack \cite{MBPTrack} introduces a memory bank module that disseminates target information across multiple historical frames into the search area. SeqTrack3D \cite{SeqTrack3D} and StreamTrack \cite{steamtrack} sequentially integrate information from multiple point cloud frames, thereby reformulating the tracking task as a sequence processing problem. In particular, SeqTrack3D \cite{SeqTrack3D} employs bounding box sequences predicted from the preliminary stage to guide the learning of point cloud sequence features, utilizing historical bounding boxes as constraints to regulate the network’s modeling of target motion patterns. StreamTrack \cite{steamtrack} adopts a hybrid attention mechanism to carefully extract the spatial and temporal information from point cloud sequences. The aforementioned methods primarily address tracking in normal settings.

\subsection{3D Tracking in Specialized Settings} 
% 上述方法聚焦于常规场景的跟踪，但是在特殊场景下的跟踪也很常见，比如远距离小尺度、恶劣天气、高时序变化、对抗攻击等, 在该种情况下以上方法的性能均有明显的退化。Tian \etal \cite{sot_size} 面向目标尺度变化，提出了一种应用原型特征的方式来突出前景点，并为小目标给出了一种特征图细分设计。Dreissig \etal \cite{lidar_weather} 则关注到了LIDAR感知算法在面对雨雪雾等天气是的性能下降表现。Tian\etal \cite{sot_attack} 研究了当前不同框架下的跟踪方法在遭受白盒攻击和黑盒攻击时的表现。Wu \etal \cite{hvtrack} 等人针对高时序变化场景，设计了一种位移感知的记忆池机制来处理大的形状变化。本文也聚焦于高时序变化场景，其为了缓解以前研究的以下局限：temporal information redundancy within the memory bank，absence of geometric prior information in cross attention.
In fact, tracking in specialized scenarios, such as distant and small objects, adverse weather conditions, high temporal variations, and adversarial attacks, is also highly prevalent. In such settings, the performance of the aforementioned methods often degrades markedly. Tian \etal \cite{ClassAgnostic_Tracking} observed that the current trackers suffer from significant degradation when conducting class-agnostic tracking, and thus proposed the generation of a common pattern throughout the feature fusion process. Similarly, the challenges posed by distant and small objects were tackled in \cite{sot_size} through the design of a prototype feature and subdivision modules to emphasize foreground targets. Dreissig \etal \cite{lidar_weather} investigated the vulnerability of the LiDAR perception algorithms in conditions of rain, snow, and fog. Tian \etal \cite{sot_attack} evaluated the robustness of three main tracking frameworks, specifically in the presence of white-box and black-box attacks. For high temporal variations, Wu \etal \cite{hvtrack} introduced a pose-aware memory bank to handle large shape changes, similar object distractions, and background noise. In this work, we also concentrate on high temporal variation tracking task, and aim to mitigate the following limitations of previous studies: insufficient computational efficiency, temporal redundancy in the memory bank, and the lack of historical geometric prior in feature extraction.

\subsection{State Space Models for Information Processing}
Mamba model \cite{mamba, mamba2}, introduced by Gu and Dao, is a state space model designed for efficient long-sequence modeling derived from the continuous system. Building on the S4 framework \cite{s4}, it introduces an input-dependent selection mechanism and a hardware-friendly selective scan, enabling efficient long-range dependency modeling with near-linear complexity. 
Initially applied to 1D sequences, Mamba has been adapting to 2D vision tasks \cite{MambaVision, mambaout}. MambaVision \cite{MambaVision} introduces a vision-friendly Mamba block within a hybrid Mamba–Transformer backbone that combines CNN-based early stages with self-attention in later layers, achieving advanced accuracy–throughput trade-offs and superior performance across classification, detection, and segmentation benchmarks.
However, applying Mamba to 3D point clouds poses challenges: (1) unordered point sets can cause pseudo-order dependence; (2) point cloud understanding relies on fine-grained local geometric structures, which are not explicitly captured by the original Mamba design; and (3) global–local feature integration is non-trivial.
Several works have explored Mamba for 3D point cloud classification and detection \cite{PointMamba, unimamba_detection, PCMamba, mamba3d}. Specifically, PointMamba \cite{PointMamba} directly models point patch sequences but lacks explicit local geometry; PCM \cite{PCMamba} integrates Mamba into PointMLP but with high computational cost; Mamba3D \cite{mamba3d} introduces Local Norm Pooling for local geometry and a bidirectional SSM to mitigate pseudo-order effects, achieving state-of-the-art results with linear complexity. Although these Mamba-based point cloud classification methods hold promise as backbones for 3D HTV tracking, they still face challenges in streaming point cloud video scenarios and lack dedicated tracking-oriented designs to address temporal redundancy in the memory bank and the incorporation of historical geometric priors. 
\section{Method}

\begin{figure*}[htbp]
\centering
\includegraphics[width=1\linewidth]{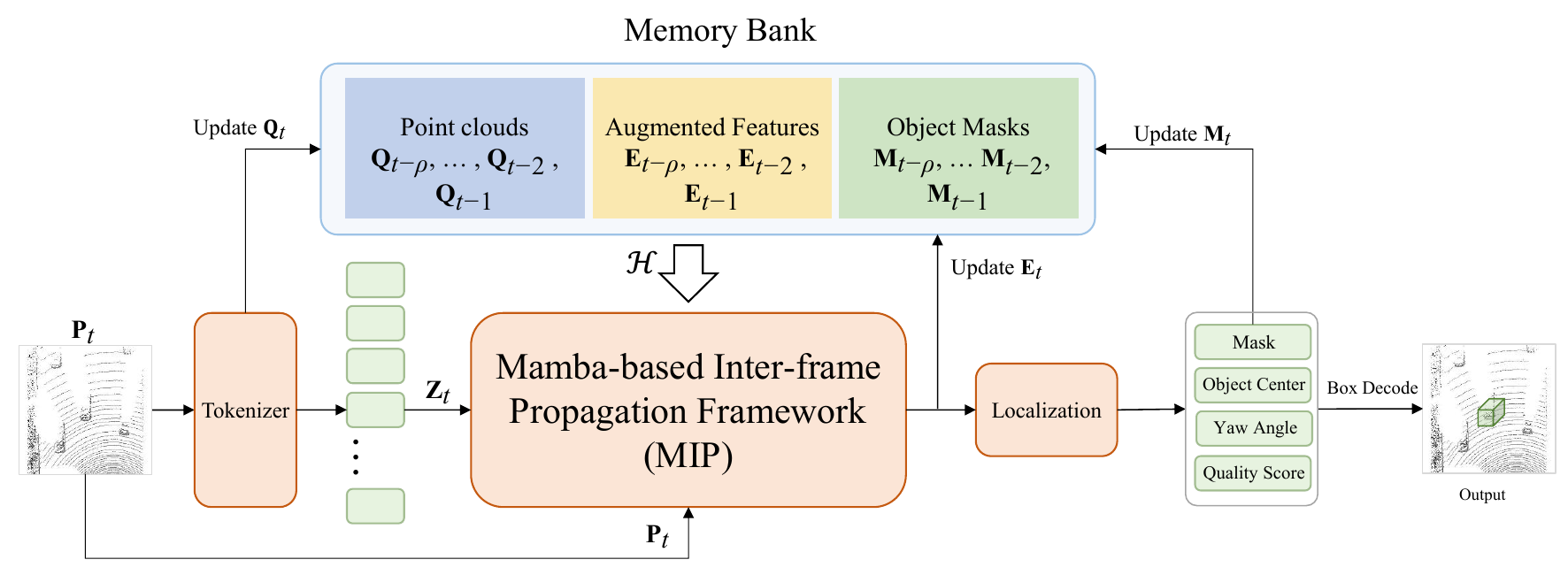}
\caption{\textbf{The overall pipeline of the proposed method.}}
\label{fig:overview}
\end{figure*}

Our method aims to deal with the three challenges that are suffered in HTV scenarios:
\begin{enumerate}
    \item[Q1:] How to improve the computational efficiency from the existing quadratic complexity;
    \item[Q2:] How to improve the tracking accuracy with consideration of spatial relationships among historical frames;
    \item[Q3:] How to mitigate temporal redundancy.
\end{enumerate}

In the following subsections, we will first formulate the LiDAR VOT into a mathematical problem. Then, we will present a Mamba-based inter-frame propagation framework, which is capable of improving run speed without sacrificing tracking accuracy. Finally, we will introduce a grouped feature enhancement module to alleviate temporal redundancy.

\subsection{Problem Definition}
To conduct visual object tracking from the LiDAR data, the task begins by identifying a specific target within a 3D bounding box in the first frame of a point cloud sequence. Each frame is represented as a point cloud $\mathbf{P}_t \in \mathbb{R}^{N_t \times 3}$, where $N_t$ denotes the number of points at time $t$. The target bounding box is parameterized by its center coordinates $(x, y, z)$, size $(w, l, h)$, and heading angle $\theta$ around the up-axis.

Formally, given an initial point cloud $\mathbf{P}_1$ and the object of interest covered by the 3D bounding box $\mathbf{B}_1$, the goal is to recursively estimate the 7-degree-of-freedom state $\mathbf{B}_t = (x_t, y_t, z_t, w_t, l_t, h_t, \theta_t)$ in each subsequent frame $\mathbf{P}_t$ for $t = 2, \dots, T$. In transportation scenarios, most elements on the road are usually rigid, thus their physical sizes $(w, l, h)$ remain unchanged in 3D space. Taking the point cloud sequences as input, the LiDAR VOT simplifies to estimating the translational offsets $(\delta^x_t, \delta^y_t, \delta^z_t)$ and the angular offset $\delta^\theta_t$ online. In this study, the proposed approach leverages sequential modeling through a deep learning-based state estimator that incorporates historical context into the current search region. The process can be formulated as:
\begin{equation}
\Phi\big( \mathbf{P}_t; \mathcal{H} \big) \mapsto (\delta^x_t, \delta^y_t, \delta^z_t, \delta^\theta_t), 
\end{equation}
where $\mathcal{H}$ encodes historical target states and environments, enabling robust and accurate tracking through temporal reasoning.

\subsection{Mamba-based Inter-frame Propagation Framework}
In this subsection, we first provide a brief overview of Mamba, and then present the detailed design of the Mamba-based Inter-frame Propagation (MIP) framework.

% Mamba用于描述系统随时间的动态行为，其状态方程可表示为$h'(t)=Ah(t) + Bx(t)$. 为了计算机编程实现，通过零阶保持法将其离散化为迭代形式：$h_{k+1}=\overline{A}h_t + \overline{B}x_k$, 其中$\overline{A} = exp(A\delta)$, $\overline{B} = A^{-1}(exp(A\delta) - I)B$, $\delta=x_{t+1}-x_t$. 考虑输出投影矩阵$\overline{C}$, 最终的输出y可以利用卷积形式来实现并行计算：$y=\mathbf{x}*K$, 其中$K=(\overline{CA^0B}, \overline{CA^1B},\dots, \overline{CA^kB})$, $\mathbf{x} = (x_0, x_1, \dots, \x_k)$. 然而，上述参数ABC固定的，无法做到输入感知，这不限制了学习过程中选择记忆的能力。因此，mamba设计了输入感知的动态权重学习机制并适配了硬件感知的选择扫面机制以实现并线计算：$h_{k+1}=\hat{A}(x_k)h_t + \hat{B}(x_k)x_k$. 详细信息请参考文献\cite{mamba}.
\textbf{Mamba Review.} State space model is used to represent the dynamic behavior of a system over time. Its state equation can be written as $h'(t) = A h(t) + B x(t)$. For implementation in computer programming, it is discretized using the Zero-Order Hold method into an iterative form: $h_{k+1} = \overline{A} h_k + \overline{B} x_k$, where $\overline{A} = \exp(A \Delta)$ and $\overline{B} = A^{-1}(\exp(A \Delta) - I) B$, and $\Delta = t_{k+1} - t_k$.  
Considering the output projection matrix $\overline{C}$, the final output $y$ can be achieved in a parallel computation manner using a convolution operation: $y = \mathbf{x} * K$,  where $K = (\overline{C} \overline{A}^0 \overline{B}, \overline{C} \overline{A}^1 \overline{B}, \dots, \overline{C} \overline{A}^k \overline{B})$ and $\mathbf{x} = (x_0, x_1, \dots, x_k)$. 

After training, however, the parameters $\overline{A}$, $\overline{B}$, and $\overline{C}$ are fixed, which prevents the model from being input-aware and limits its ability to selectively retain attention. Therefore, Mamba introduces an input-dependent dynamic weight learning mechanism and incorporates a hardware-aware selective scanning algorithm \cite{mamba} to enable parallel computation: $h_{k+1} = \hat{A}(x_k) h_k + \hat{B}(x_k) x_k$. Please refer to \cite{mamba} for more details.

% 尽管Transformer架构已经在该任务上获得令人印象深刻的结果，但是它也因二次复杂度的缺点影响着实时场景的应用。图1中展示了随着输入点云数据的增加，基于Transformer架构方法（如MBPTrack）其运行速度和计算量显著增加。鉴于此，为了在高时态变化场景中实现高效计算, 我们创新性地采用状态空间模型来处理LiDAR跟踪问题；为了提高跟踪的精度，我们设计了一种借助空间关系和时序历史信息的特征提传播方式以替代传统的单帧特征提取。将这两个部分进行有机耦合就是本文Mamba-based Inter-frame Propagation Framework （MIP）的主要动机。接下来，我们先简要回顾状态空间模型Mamba，然后给出MIP的具体介绍。
\begin{figure}[htbp]
\centering
\includegraphics[width=1\linewidth]{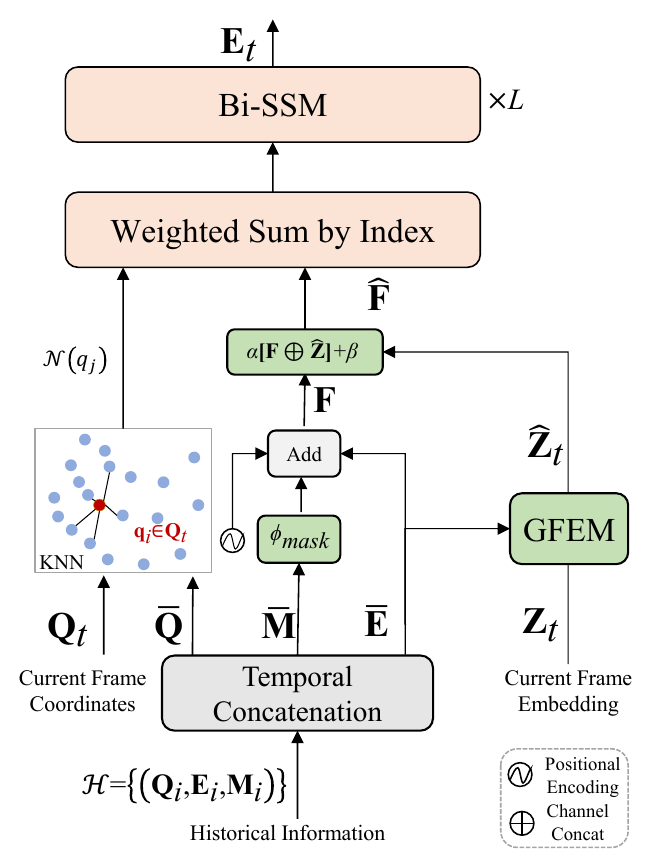}
\caption{\textbf{The pipeline of the proposed MIP module.}}
\label{fig:mip}
\end{figure}
\textbf{The MIP.} Although Transformer-based architectures have achieved impressive results on the LiDAR VOT, their quadratic complexity poses a significant limitation for real-time applications. As illustrated in Fig. \ref{fig:flops}, when the input quantity of point clouds increases, the Transformer-based methods (\eg, MBPTrack \cite{MBPTrack}) suffer from a sharp rise in both computational cost and runtime latency. 
To enable efficient computation listed in the Q1, we deal with the LiDAR VOT based on the state space model, Mamba, which excels at near-linear complexity. Furthermore, to enhance tracking accuracy stated in the Q2, we design a feature propagation mechanism that exploits both spatial relationships and temporal historical information, in contrast to conventional single-frame feature extraction. The integration of these two ideas constitutes the core motivation behind our MIP. 

% 图给出了MIP的帧间传播框架。MIP的输入包括两个部分：待处理的当前帧$\mathbf{P}_t$以及记忆池$\mathcal{H}$。一方面，对于$\mathbf{P}_t$， 我们首先利用最远点采样获取关键点，并每个关键点的K邻域内提取局部特征。另一方面，为了充分利用历史信息，记忆池由点云坐标$\mathbf{P}$、特征$\mathbf{E}$以及目标掩码$\mathbf{M}$三个基本元素组合成，\ie, $mathcal{H} = \{(\mathbf{P}_i, \mathbf{E}_i, \mathbf{M}_i)| \mathbf{P}_i\in \mathbb{R}^{N_t \times 3}, \mathbf{E}_i\in \mathbb{R}^{N_i \times C},\mathbf{M}_i\in \mathbb{R}^{N_t \times 1}, i=1,2, \dots, t-1\}_t$. 
The figure \ref{fig:mip} illustrates the MIP framework. The input consists of two components: the current frame $\mathbf{P}_t$ and the memory bank $\mathcal{H}$. On one hand, as the input of the state space model, ``tokens'' are first obtained from the point cloud:
\begin{equation}
    \mathbf{Z}_t = \mathcal{K}(\mathcal{S}(\mathbf{P}_t)),
\end{equation}
where $\mathcal{S}$ means farthest point sampling and $\mathcal{K}$ utilizes mini-PointNet \cite{PointNet2017} to learn the local feature of each sampled point within its K-Nearest Neighborhood (KNN). On the other hand, to fully leverage temporal historical cues, the memory bank $\mathcal{H}$ is spanned by three basic elements: point cloud coordinates $\mathbf{Q_i}=\mathcal{S}(\mathbf{P}_i)$, features $\mathbf{E}_i$, and target masks $\mathbf{M}_i$. It can be written as $\mathcal{H} = \{(\mathbf{Q}_i, \mathbf{E}_i, \mathbf{M}_i) \mid \mathbf{Q}_i \in \mathbb{R}^{\hat{N} \times 3}, \mathbf{E}_i \in \mathbb{R}^{\hat{N} \times C}, \mathbf{M}_i \in \mathbb{R}^{\hat{N} \times 1}, i = t-\rho, t-\rho+1, \dots, t-1\}$. Herein, $\rho$ represents the number of past frames stored in the memory bank. 

Based on the above inputs, we first concatenate the historical frames in temporal order:  
\begin{align}
    \mathbf{\bar Q} &= [\mathbf{Q}_{t-\sigma}; \dots; \mathbf{Q}_{t-2}; \mathbf{Q}_{t-1}],\\
    \mathbf{\bar E} &= [\mathbf{E}_{t-\sigma}; \dots; \mathbf{E}_{t-2}; \mathbf{E}_{t-1}],\\
    \mathbf{\bar M} &= [\mathbf{M}_{t-\sigma}; \dots; \mathbf{M}_{t-2}; \mathbf{M}_{t-1}].
\end{align}

The goal is to exploit historical cues to obtain more discriminative target features for the current frame \(\mathbf{Q}_t\). Prior to feature propagation, the mask features \(\mathbf{\bar M}\) are integrated into the appearance features \(\mathbf{\bar E}\):  
\begin{equation}
    \mathbf{F} = \mathbf{\bar{E}} + \phi_{mask}(\mathbf{\bar{M}}),
\end{equation}
where \(\phi_{mask}:\mathbb{R}^{\rho\hat{N}\times 1} \mapsto \mathbb{R}^{\rho\hat{N}\times C}\) is a 1D convolution that learns mask-aware representations.  

For each point \(q_j \in \mathbf{Q}_t\), we construct its local neighborhood \(\mathcal{N}(q_j)\) in the aggregated historical point cloud \(\mathbf{\bar Q}\) using KNN. The corresponding historical features \(\{\mathbf{F}_k \mid k \in \mathcal{N}(q_j)\}\) are then retrieved from \(\mathbf{F}\). The propagated feature of \(q_j\) is computed as:  
\begin{align}
    \mathbf{e}_{t,j} &= \sum_{k\in\mathcal{N}(q_j)} \mathbf{\omega}_k \odot \hat{\mathbf{F}}_k,\\
    \hat{\mathbf{F}}_k &= \alpha[\mathbf{F}_k \oplus \mathbf{Z}_{t}(q_j)] + \beta,\\ 
    \mathbf{\omega}_k &= \frac{\exp(\hat{\mathbf{F}}_k)}{\sum_{l\in\mathcal{N}(q_j)} \exp(\hat{\mathbf{F}}_l)},\\
    \mathbf{E}_t &= [\mathbf{e}_{t,1}, \mathbf{e}_{t,2}, \dots, \mathbf{e}_{t,\hat N}],
\end{align}
where \(\odot\) denotes element-wise multiplication, \(\oplus\) denotes channel concatenation, \(\mathbf{z}_{t}(q_j)\in \mathbb{R}^{C}\) is the initial token embedding of the current point, and \(\alpha,\beta\) are learnable parameters. 

Finally, the propagated features are fed into Bi-SSM \cite{mamba3d}, which effectively handles unordered and irregular point sets:  
\begin{equation}
     h_{k+1} = \hat{A}(\mathbf{e}_{t,j}) h_k + \hat{B}(\mathbf{e}_{t,j}) \mathbf{e}_{t,j}.
\end{equation}

% % 基于上述的输入信息，我们先将历史帧按时间顺序进行拼接融合：
% \begin{align}
%     \mathbf{\bar Q} &= [\mathbf{Q}_{t-\sigma}; \dots; \mathbf{Q}_{t-2}; \mathbf{Q}_{t-1}],\\
%     \mathbf{\bar E} &= [\mathbf{E}_{t-\sigma}; \dots; \mathbf{E}_{t-2}; \mathbf{E}_{t-1}],\\
%     \mathbf{\bar M} &= [\mathbf{M}_{t-\sigma}; \dots; \mathbf{M}_{t-2}; \mathbf{M}_{t-1}].
% \end{align}
% % 此处的目标是利用历史帧提供的线索，为当前帧点云$\mathbf{Q}_t$获得更具有判别性目标特征表示。因此，在进行特征传播之前，我们将掩码特征\mathbf{\bar M}z整合进外观特征$\mathbf{\bar E}$之中：
% \begin{equation}
%     \mathbf{F} = \mathbf{\bar{E}} + \phi_{mask}(\mathbf{\bar{M}}),
% \end{equation}
% % 其中$\phi_{mask}:\mathbb{R}^{\sigma\hat{N}\times 1} \mapsto \mathbb{R}^{\rho\hat{N}\times C}$为一维卷积旨在去学习掩码特征。
% %随后，对于$\mathbf{Q}_t$中的每一点$q_j$，我们利用KNN在历史帧聚合点云$\mathbf{\bar Q}$中构建其局部几何邻域$\mathcal{N}(q_j)$. 根据此邻域索引，我们在融合特征$\mathbf{F}$中查找对应的历史特征$\{ \mathbf{F}_k|k \in \mathcal{N}(q_j), \mathbf{F}_k \in \mathbb{R}^{C} \}$。每一点$q_j$的特征可以表示为邻域的加权求和：
% \begin{align}
%     \mathbf{e}_{t,j} &= \sum_{k\in\mathcal{N}(q_j)} \mathbf{\omega}_k \odot \hat{\mathbf{F}}_k,\\
%     \hat{\mathbf{F}}_k &= \alpha[\mathbf{F}_k \oplus \mathbf{E}_{t}(q_j)] + \beta,\\ 
%     \mathbf{\omega}_k &= \frac{\exp(\hat{\mathbf{F}}_k)}{\sum_{l\in\mathcal{N}(q_j)} \exp(\hat{\mathbf{F}}_l)}, \\
% \end{align}
% % 其中，$\odot$表示矩阵元素相乘，$\oplus$表示按通道拼接，$\mathbf{E}_{t}(q_j)\in \mathbb{R}^{C}$为当前点初始token embedding，$\alpha$和$\beta$为可学习参数。最终，将此传播特征被送入Bi-SMM\cite{mamba3d}，which is capable of dealing with unordered and irregular for point clouds, 即：
%  \begin{equation}
%      h_{k+1} = \hat{A}(\mathbf{e}_{t,j}) h_k + \hat{B}(\mathbf{e}_{t,j}) \mathbf{e}_{t,j}
%  \end{equation}
 
\subsection{Grouped Feature Enhancement Module}
% 由上述公式（8）中可知，当前帧的点云特征主要基于历史帧特征的传播方式产生。但是，记忆池中的冗余特征会使得在邻域中计算的特征缺少判别性。鉴于此，我们从特征通道入手来增强前景和背景表示。The underlying motivation is that each feature channel should be dedicated to a single category of points (either foreground or background) \cite{ClassAgnostic_Tracking}. Accordingly, we divide the features into two groups and apply distinct linear transformations to process foreground and background points separately.

% 具体地，对于给定的$\mathbf{E}_t\in \mathbb{R}^{\hat{N}\times C}$以及$\mathbf{\bar E}\in \mathbb{R}^{\rho\hat{N}\times C}$, 我们将其按照通道分为两组：$\mathbf{E}_t^1,\mathbf{E}_t^2\in \mathbb{R}^{\hat{N}\times \frac{C}{2}}$; $\mathbf{\bar E}^1, \mathbf{\bar E}^2\in \mathbb{R}^{\rho\hat{N}\times \frac{C}{2}}$. 对每一组，进行交叉注意力计算：
% \begin{equation}
%     \mathbf{\hat E}_t^1 = (\mathbf{E}_t^1 \mathbf{W}_Q^1) (\mathbf{\bar E}^1 \mathbf{W}_K^1)^\top (\mathbf{\bar E}^1 \mathbf{W}_V^1),
% \end{equation}
% 其中$\mathbf{W}_Q^1, \mathbf{W}_K^1, \mathbf{W}_V^1$为可学习的参数矩阵。同理， $\mathbf{\hat E}_t^2$也可以此种方式计算得到。将二者按通道拼接可得到$\mathbf{\hat E}_t = \mathbf{\hat E}_t^1 \oplus \mathbf{\hat E}_t^2$.最终，公式（8）变为：
% \begin{equation}
%      \hat{\mathbf{F}}_k = \alpha[\mathbf{F}_k \oplus \mathbf{\hat E}_{t}(q_j)] + \beta.
% \end{equation}
% The resulting features can facilitate distinguishing the correlated points between the current frame and the memory bank, further alleviating temporal redundancy caused by aggregating historical frames.
In fact, the redundancy stated in the Q3 can be alleviated by constructing such local neighborhoods in historical frames and selecting top-K keypoints. However, redundant information stored in the memory bank may reduce the discriminative power of the features computed within local neighborhoods. To further address the Q3, we enhance the representation of foreground and background points from the perspective of feature channels.
Our underlying motivation is that each feature channel should be dedicated to a single category of points (either foreground or background) \cite{ClassAgnostic_Tracking}. Specifically, based on Eq. (8), it can be observed that point cloud features of the current frame are primarily generated through propagation from historical frame features. We divide the features into two groups and apply distinct linear transformations to process foreground and background points separately.  

Specifically, given \(\mathbf{E}_t\in \mathbb{R}^{\hat{N}\times C}\) and \(\mathbf{\bar E}\in \mathbb{R}^{\rho\hat{N}\times C}\), we split them into two channel groups: \(\mathbf{Z}_t^1,\mathbf{Z}_t^2\in \mathbb{R}^{\hat{N}\times \frac{C}{2}}\) and \(\mathbf{\bar E}^1, \mathbf{\bar E}^2\in \mathbb{R}^{\rho\hat{N}\times \frac{C}{2}}\). For each group, cross-attention is performed as follows:  
\begin{equation}
    \mathbf{\hat Z}_t^1 = softmax[(\mathbf{Z}_t^1 \mathbf{W}_Q^1) (\mathbf{\bar E}^1 \mathbf{W}_K^1)^\top] (\mathbf{\bar E}^1 \mathbf{W}_V^1),
\end{equation}
where \(\mathbf{W}_Q^1, \mathbf{W}_K^1, \mathbf{W}_V^1\) are learnable parameter matrices. Similarly, \(\mathbf{\hat E}_t^2\) can be computed in the same manner. Concatenating the two groups along the channel dimension yields \(\mathbf{E}_t = \mathbf{\hat Z}_t^1 \oplus \mathbf{\hat Z}_t^2\). Consequently, Eq. (8) is reformulated as:  
\begin{equation}
     \hat{\mathbf{F}}_k = \alpha[\mathbf{F}_k \oplus \mathbf{\hat Z}_{t}(q_j)] + \beta.
\end{equation}
These enhanced representations improve the discrimination of correlated points across the current frame and the memory bank, effectively reducing the temporal redundancy accumulated from historical frame aggregation. 

\begin{figure}[tbp]
\centering
\includegraphics[width=0.8\linewidth]{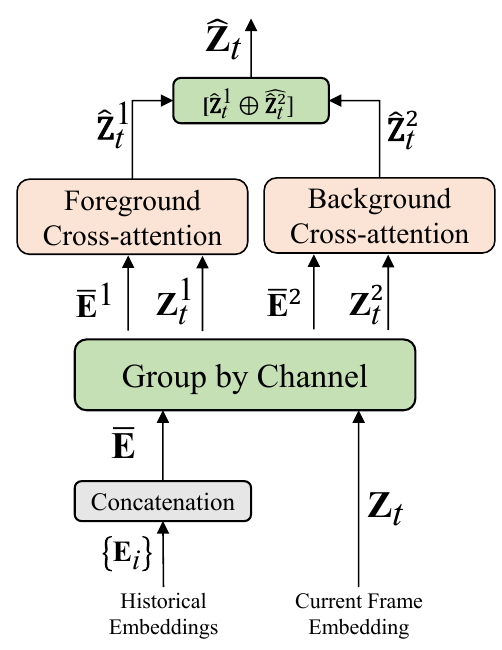}
\caption{\textbf{The pipeline of the proposed GFEM module.}}
\label{fig:GFEM}
\end{figure}

\subsection{Implementation Details}
% 由MIP和GFEM模块处理后的特征最终送入基于包围盒先验的定位网络\cite{MBPTrack}, 该部分的输出包括当前帧前背景分类结果、hough voting的中心预测及其质量分数、目标包围盒的参数及其质量分数。损失函数同样采用MBPTrack中的形式：$L = L_m + L_c + L_q + L_s + L_b$， 其中$L_m$为交叉熵损失来约束前背景掩码的生成，$L_c$为均方误差损失来约束hough voting的中心预测，$L_q$和$L_s$为交叉熵损失来分别约束hough voting 的质量分数和包围盒的质量分数，$L_b$为smooth-L1损失来约束包围盒的参数。

% 为了在帧间传播进行训练，我们在LiDAR点云序列中以8帧为单位采样目标片段，每个片段为一个训练样本。在生成token时，我们将每一帧的token数量设置为$\hat{N}=128$，特征的维度$C=128$。记忆池样本数量设置为$\rho=3$。邻域集合\(\mathcal{N}(q_j)\)的邻居点数量设置为$4$.

The features processed by the MIP and GFEM modules are finally fed into a box-Prior localization network \cite{MBPTrack}. The outputs of this network include: foreground–background classification results for the current frame, Hough voting–based center predictions and their quality scores, and the parameters of the 3D bounding box along with their quality scores. The loss function follows the same formulation as in MBPTrack:  
\begin{equation}
    \mathcal{L} = \mathcal{L}_m + \mathcal{L}_c + \mathcal{L}_q + \mathcal{L}_s + \mathcal{L}_b,
\end{equation}
where \(\mathcal{L}_m\) is a cross-entropy loss constraining the generation of the foreground–background mask, \(\mathcal{L}_c\) is a mean squared error loss supervising the Hough voting center prediction, \(\mathcal{L}_q\) and \(\mathcal{L}_s\) are cross-entropy losses supervising the quality scores of the Hough voting and bounding box respectively, and \(\mathcal{L}_b\) is a smooth-L1 loss constraining the bounding box parameters.  

For inter-frame propagation training, we sample target segments of 8 consecutive frames from LiDAR point cloud sequences, with each segment serving as a training instance. During token generation, the number of tokens per frame is set to \(\hat{N}=128\), and the feature dimension is set to \(C=128\). The memory bank size is fixed at \(\rho=3\). For each point \(q_j\), the neighborhood set \(\mathcal{N}(q_j)\) is constructed with 4 nearest neighbors.  
\section{Experiments}

\subsection{Datasets and Metrics}
\textbf{Datasets.} 
% 本文采用面向室外场景的由激光雷达采集到的点云数据开展研究。KITTI和nuscenes是当前代表性的大规模室外场景数据集，其提供了物体实例随时间变化的三维包围盒标注，可用于目标跟踪的训练和测试。KITTI是由一个64线的激光雷达采集获得，其每秒可采集超过130万个点，频率为10HZ。我们采用其发布的KITTI Tracking数据集作为实验对象，其由21个场景组成，包括道路场景、住宅区、校园等。该数据集进行了逐帧的标注，为了检验算法在应对高时序变化的能力，我们采用文献1的方式，对每个跟踪实例进行有间隔的帧间采样。为了全面衡量在不同时序变化下的能力，其采样间隔设置为2，3，5，和10。nuScenen是由32线激光雷达采集得到，频率为20HZ，每秒可达140万个点。该数据集记录了1000个场景，每个场景持续20秒左右。由于数据标注耗时耗力，nuscene数据只对关键帧进行了标注，即其每隔0.5秒标记一帧。这两种传感器的差异也会更全面考验算法的适用性。此外，我们还在另一个规模更大的数据集上检验算法的性能：Waymo Open Set，该数据集利用频率为10HZ的LiDAR设备采集了1150个场景，包括城市道路、高速公路以及不同的天气场景。本文基于文献2和3，对waymo数据集进行二次加工，给出了面向高时序变化的跟踪数据平台。其将每个跟踪目标分为简单、中等和困难三个等级，并按照时间间隔2，3，5，10对点云序列进行采样。最后的数据集包含汽车总帧数为185632，行人总帧数为241168.
In this study, we conduct experiments on outdoor point cloud data collected by LiDAR sensors. KITTI \cite{KITTI2013} and nuScenes \cite{nuscenes} are two representative large-scale outdoor datasets that provide 3D bounding box annotations of object instances over time, making them widely used benchmarks for training and evaluating tracking algorithms.  

The KITTI dataset was collected using a 64-beam LiDAR sensor, capable of capturing more than 1.3 million points per second at a frequency of 10 Hz. We adopt the KITTI Tracking benchmark, which consists of 21 scenes covering diverse environments such as roadways, residential areas, and campuses. This dataset provides frame-by-frame annotations. To evaluate the robustness of our method under high temporal variation, we follow the sampling strategy in \cite{hvtrack}, where each tracking sequence is subsampled at different frame intervals. Specifically, we set the sampling intervals to 2, 3, 5, and 10 frames to comprehensively assess performance under varying temporal gaps.  

The nuScenes dataset was collected using a 32-beam LiDAR operating at 20 Hz, capturing up to 1.4 million points per second. It contains 1000 driving scenes, each lasting approximately 20 seconds. Due to the high cost of annotation, only key frames are labeled, with one frame annotated every 0.5 seconds. The differences between these two LiDAR sensors (64-beam vs. 32-beam, dense vs. sparse sampling) provide a more comprehensive evaluation of the adaptability and robustness of the proposed tracking algorithm.  

\textbf{Evaluation metrics.} For the evaluation metrics, the success and precision rates were used, following the framework of a one-pass evaluation \cite{sot_eval}. The success rate is represented in the area under the curve (AUC) plot, where the X-axis denotes the intersection over union (IOU) between the predicted and ground-truth bounding boxes, and the Y-axis indicates the proportion of frames exceeding the specified threshold. The IOU value is measured on a continuous scale from 0 to 1. Conversely, the precision is also calculated through an AUC plot; in this case, the X-axis measures the center errors between the predictions and ground-truth, while the Y-axis quantifies those frames where the center error falls below a certain threshold. 

\subsection{Main Results on High Temporal Variation}
% 当前LIDAR点云跟踪的方法可根据处理场景的不同划分为两类：面向常规场景的跟踪和面向特殊场景的跟踪。本文聚焦于后者，详细的描述即为面向高时序变化的点云场景跟踪。HVTrack是此类任务的先行者，本文则将其作为基线模型，旨在通过时序信息和几何信息的高效利用来提升跟踪精度。除此之外，为了全面的评测方法的有效性，我们也在高时序变化的实验设置下对比了面向常规场景的前沿跟踪方法，包括 P2B\cite{P2B},  BAT\cite{BAT}, M2Track\cite{MMTrack}, CXTrack\cite{cxtrack}, M3SOT\cite{m3sot}以及MBPTrack\cite{MBPTrack}. 下面分别从KITTI和nuScenes数据集对结果进行分析。
Current LiDAR-based tracking methods can be divided into two categories: those designed for normal scenarios and those tailored for specialized scenarios. This work focuses on the latter, specifically high temporal variation (HTV) tracking. HVTrack is the pioneering baseline for this task, and we adopt it as our reference model, aiming to improve accuracy through efficient utilization of temporal and geometric information. To comprehensively validate our approach, denoted as MambaTrack, we also evaluate it under HTV settings against state-of-the-art trackers originally developed for normal scenarios, including P2B \cite{P2B}, BAT \cite{BAT}, M2Track \cite{MMTrack}, CXTrack \cite{cxtrack}, M3SOT \cite{m3sot}, and MBPTrack \cite{MBPTrack}. Detailed results on the KITTI and nuScenes datasets are analyzed in the following subsections.

\renewcommand{\arraystretch}{1.5} %line height
\begin{table*}[ht]
  \centering
  \caption{\textbf{Comparison of MambaTrack3D with the state-of-the-art methods on each category of the KITTI-HTV dataset}. The KITTI-HTV dataset is constructed for training and testing by sampling different frame intervals in the KITTI dataset. \textbf{Bold} denotes the best results. ``Success/Precision'' are used for evaluation. Improvement and deterioration are shown in \green{green} and \red{red}, respectively.}
  \resizebox{\linewidth}{!}{
  \begin{tabular}{c|ccccc|ccccc}
    \toprule[1.5pt]
            & \multicolumn{5}{c|}{2 Intervals}      & \multicolumn{5}{c}{3 Intervals} \\
    \midrule
    Methods & Car   & Pedestrian & Van   & Cyclist & \textbf{Mean}  & Car   & Pedestrian & Van   & Cyclist & \textbf{Mean} \\
    (Frame Number) & (6424)  & (6088)  & (1248)  & (308)   & (14068) & (6424)  & (6088)  & (1248)  & (308)   & (14068) \\
    \midrule
    P2B\cite{P2B}   & 56.3/71.0 & 30.8/53.0 & 33.4/38.4 & 41.8/61.4 & 42.9/60.1 & 43.4/51.8 & 27.9/46.8 & 27.9/31.8 & 44.8/64.4 & 35.4/48.1 \\
    BAT~\cite{BAT}   & 61.8/74.2 & 36.5/61.1 & 26.8/30.4 & 54.1/78.7 & 47.6/64.7 & 51.7/61.9 & 31.8/53.5 & 24.0/28.2 & 50.5/72.6 & 40.6/55.5 \\
    M2Track~\cite{MMTrack} & 63.0/76.6 & 54.6/81.7 & 52.8/66.5 & 68.3/89.3 & 58.6/78.2 & 62.1/72.7 & 51.8/74.3 & 33.6/41.6 & 64.7/82.0 & 55.1/70.8 \\
    CXTrack~\cite{cxtrack} & 61.4/70.9 & 62.6/86.3 & 56.0/69.1 & 59.2/76.9 & 61.4/77.5 & 47.4/53.1 & 57.9/79.3 & 48.5/58.8 & 40.7/58.4 & 51.9/65.1 \\
    M3SOT~\cite{m3sot}   & 59.0/67.9 & 61.7/86.3  & 55.2/68.7 & 55.1/86.3 & 59.8/76.3 & 46.9/52.6 & 50.1/74.0 & 43.3/53.7 & 32.4/48.1 & 47.7/61.9 \\
    MBPTrack~\cite{MBPTrack} & 70.3/80.2 & 58.6/81.5 & 50.7/59.7 & 65.5/82.0 & 63.4/79.0 & 64.3/73.3 & 48.2/69.7 & 43.6/50.0 & 68.6/86.6 & 55.6/70.0 \\
    \midrule
    HVTrack~\cite{hvtrack} & 67.1/77.5 & 60.0/84.0 & 50.6/61.7 & 73.9/93.6 & 62.7/79.3 & 66.8/76.5 & 51.1/71.9 & 38.7/46.9 & 66.5/89.7 & 57.5/72.2 \\
    Ours                  & 70.0/81.9 & 64.3/89.5 & 64.2/76.2 & 77.3/94.8 & \textbf{67.2/85.0} & 69.5/80.9 & 59.5/85.6 & 56.8/65.7 & 69.6/88.1 & \textbf{64.0/81.7} \\
                          &   &   &  &  & \green{$\uparrow 4.5/\uparrow 5.7$}  & & & & & \green{$\uparrow 6.5/\uparrow 9.5$}\\
    \midrule
    \midrule
             & \multicolumn{5}{c|}{5 Intervals}      & \multicolumn{5}{c}{10 Intervals} \\
    \midrule
    Methods & Car   & Pedestrian & Van   & Cyclist & \textbf{Mean}  & Car   & Pedestrian & Van   & Cyclist & \textbf{Mean} \\
    (Frame Number) & (6424)  & (6088)  & (1248)  & (308)   & (14068) & (6424)  & (6088)  & (1248)  & (308)   & (14068) \\
    \midrule
    P2B~\cite{P2B}   & 39.3/46.1 & 27.4/43.5 & 27.2/30.4 & 35.0/44.4 & 33.0/43.5 & 28.6/29.2 & 23.1/31.1 & 25.9/27.3 & 29.1/28.3 & 26.0/29.8 \\
    BAT~\cite{BAT}   & 44.1/51.1 & 21.1/32.8 & 26.1/29.5 & 35.7/46.3 & 32.4/41.1 & 30.6/33.1 & 21.7/29.2 & 20.8/20.7 & 29.3/29.1 & 25.9/30.2 \\
    M2Track~\cite{MMTrack} & 50.9/58.6 & 31.6/45.4 & 30.0/36.5 & 47.4/61.0 & 40.6/51.0 & 33.0/35.1 & 17.5/24.1 & 20.7/20.8 & 27.7/26.6 & 25.0/28.9 \\
    CXTrack~\cite{cxtrack} & 38.6/42.2 & 35.0/47.8 & 21.6/24.3 & 25.7/33.3 & 35.3/42.8 & 30.2/32.4 & 18.2/21.4 & 17.5/17.9 & 27.7/26.5 & 23.8/26.2 \\
    M3SOT\cite{m3sot}  & 30.5/34.5 & 31.0/44.0 & 18.3/21.0 & 21.6/25.9 & 29.4/37.2 & 26.1/26.6 & 16.2/18.8 & 17.6/17.1 & 27.5/26.2 & 21.1/22.4 \\
    MBPTrack\cite{MBPTrack} & 54.9/61.7 & 37.6/56.3 & 23.7/25.3 & 36.9/43.8 & 44.3/55.7 & 32.1/33.3 & 22.3/28.5 & 18.5/17.3 & 27.5/26.2 & 26.6/29.6 \\
    \midrule
    HVTrack \cite{hvtrack} & 60.3/68.9 & 35.1/52.1 & 28.7/32.4 & 58.2/71.7 & 46.6/58.5 & 49.4/54.7 & 22.5/29.1 & 22.2/23.4 & 39.5/45.4 & 35.1/\textbf{40.6} \\
    Ours                   & 62.7/72.7 & 39.5/60.1 & 49.6/55.5 & 47.4/59.6 & \textbf{51.2/65.4} & 49.1/54.3 & 22.1/28.1 & 28.8/29.6 & 34.7/38.7 & \textbf{35.3}/40.4 
    \\
                          &   &   &  &  & \green{$\uparrow 4.6/ \uparrow 6.9$}  & & & & & \green{$\uparrow 0.2 /$}\red{$\downarrow 0.2$}\\
    \bottomrule[1.5pt]
    \end{tabular}%
    }
  \label{tab:KITTI_HV}%
\end{table*}%

\textbf{Results on KITTI-HTV.}
At 2-frame intervals, MambaTrack3D achieves the best mean performance of 67.2/85.0, improving over HVTrack by +4.5/+5.7. This indicates that our framework effectively leverages temporal and geometric cues to enhance discriminability when temporal gaps are small.
At 3-frame intervals, the advantage becomes even more pronounced: MambaTrack3D reaches 64.0/81.7, outperforming HVTrack by +6.5/+9.5. This demonstrates strong robustness to moderate temporal variation, where conventional trackers degrade more sharply. 
As the temporal gap widens, performance drops for all methods at 5-frame intervals. However, MambaTrack3D still achieves the best mean score of 51.2/65.4, surpassing HVTrack by +4.6/+6.9. 
At the most challenging 10-frame interval, both methods experience significant degradation. The improvement is subtle, demonstrating that both trackers struggle under extreme temporal gaps, but MambaTrack3D maintains competitive stability.  
We observe that a similar distribution of training samples can facilitate tracking performance. For example, when using the model trained on cars to track vans, the improvement is particularly significant, with MambaTrack3D achieving 49.6/55.5 compared to 28.7/32.4 for HVTrack. These results highlight the advantages of HTV-oriented designs in maintaining robustness under large temporal gaps and demonstrate the cross-category adaptability of MambaTrack3D.

% 此外，我们从跟踪方法种类划分的视角给出以下实验发现：1）无论是面向常规场景的跟踪方法还是面向时序变化的跟踪方法,其随着帧间隔的增加性能会发生显著降低。但是，前者由于缺乏针对性的设计，衰减速率更大。
% 2) High temporal variation trackers (e.g., OTTrack and HVTrack) consistently outperform normal scenario-orientated methods (e.g., BAT and M2Track) across all intervals. 例如，时序变化的跟踪方法MambaTrack3D在间隔5帧时的表现（62.7/72.7）能媲美常规跟踪方法M3SOT在间隔仅为2时的表现（59.0%/67.9%）；
In addition, from the perspective of tracker categories, we further summarize the following findings:
\begin{itemize}
    \item Both normal-scenario trackers and high temporal variation (HTV) trackers exhibit performance degradation as the frame interval increases. However, the degradation is more severe for normal-scenario trackers due to the lack of dedicated designs for temporal variation.
    \item HTV-oriented trackers (e.g., MambaTrack3D and HVTrack) consistently outperform normal-scenario trackers (e.g., BAT and M2Track) across all intervals. For instance, MambaTrack3D achieves 62.7/72.7 at a 5-frame interval, which is comparable to the performance of the normal-scenario tracker M3SOT at only a 2-frame interval (59.0/67.9).
\end{itemize}

\begin{table*}[htbp]
  \centering 
  \caption{Comparison of MambaTrack3D with the state-of-the-art methods on each category of the NuScenes dataset. Improvement and deterioration are shown in \green{green} and \red{red}, respectively.}
  {
    \begin{tabular}{c|c|c|c|c|c|c|c}
    \toprule[1.5pt]
    \multirow{2}{*}{Type of Trackers}& Methods & Car   & Pedestrian & Truck & Trailer & Bus   & Mean \\
    & (Frame Number) & (64159) & (33227) & (13587) & (3352)  & (2953)  & (117278) \\
    \midrule
    \multirow{5}{*}{Normal Tracking}&SC3D~\cite{Siam3D2019}  & 22.3/21.9 & 11.3/12.7 & 30.7/27.7 & 35.3/28.1 & 29.4/24.1 & 20.7/20.2 \\
    &P2B~\cite{P2B}          & 38.8/43.2 & 28.4/52.2 & 43.0/41.6 & 49.0/40.1 & 33.0/27.4 & 36.5/45.1 \\
    &BAT~\cite{BAT}          & 40.7/44.3 & 28.8/53.3 & 45.3/42.6 & 52.6/44.9 & 35.4/28.0 & 38.1/45.7 \\
    &M2Track~\cite{MMTrack}  & 55.9/65.1 & 32.1/60.9 & 57.4/59.5 & 57.6/58.3 & 51.4/51.4 & 49.2/62.7 \\
    &CXTrack~\cite{cxtrack}  & 44.6/50.5 & 31.5/55.8 & 51.3/50.7 & 59.7/53.6 & 42.6/37.3 & 42.0/51.8 \\
    &MBPTrack~\cite{MBPTrack}  & 62.4/70.4 & 45.3/74.0 & 62.2/63.3 & 65.1/61.3 & 55.4/51.8 & 57.4/69.8 \\
    \midrule
    \multirow{3}{*}{Temporal variation Tracking}&HVTrack \cite{hvtrack}  & 55.9/62.9 & 41.3/67.6 & 55.6/55.2 & 52.0/40.2 & 36.3/41.6 & 51.1/62.2 \\
    &Ours                    &\textbf{64.9/73.8} & \textbf{47.1/76.2} & \textbf{64.2/66.6} & \textbf{65.7/60.4} & \textbf{59.8/56.5} &  \textbf{59.6/72.8}\\
    &             &  &  &  &  &  & \green{$\uparrow 8.5/ \uparrow 10.6$} \\
    \bottomrule[1.5pt]
    \end{tabular}%
    }
  \label{tab:nuscenes}% 
\end{table*}%

\textbf{Results on nuScene-HTV.}
The nuScenes dataset contains the tracklets where objects are annotated with bounding boxes per 10 frame intervals. Notably, its inherent sparsity (32-beam vs 64-beam) amplifies tracking challenges, particularly for small objects like pedestrians. 
MambaTrack3D achieves the best mean success/precision of 59.6/72.8, outperforming HVTrack by +8.5/+10.6. This substantial margin demonstrates our propagation and grouped feature enhancement modules effectively capture large-scale structural cues. 
Despite being small and sparse, MambaTrack3D achieves 47.1/76.2 in  pedestrians, improving over HVTrack (41.3/67.6). This indicates stronger adaptability to deformable and low-density targets. The largest improvement is observed in the bus category, with MambaTrack3D reaching 59.8/56.5 compared to HVTrack’s 36.3/41.6. This demonstrates the advantage of incorporating geometric priors for objects with large intra-class variations. 
In addition, MambaTrack3D consistently surpasses all normal-scenario trackers (e.g., MBPTrack, M2Track, CXTrack) across every category.
For example, compared to MBPTrack, which is the strongest among normal trackers (57.4/69.8), MambaTrack3D still improves by +2.2/+3.0 in mean performance. This highlights that designs tailored for high temporal variation are crucial for robust tracking in sparse LiDAR settings.

\subsection{Efficiency Comparison}
To further evaluate the computational efficiency of the proposed MambaTrack3D, we compare its FLOPs and FPS with the representative Transformer-based tracker MBPTrack under different numbers of input points, as shown in Fig. \ref{fig:flops}.  

\textbf{Runtime speed (FPS).} 
Across all input sizes, MambaTrack3D consistently achieves higher FPS than MBPTrack. For example, when the number of input points reaches 8192, MambaTrack3D still maintains real-time performance at 100 FPS, whereas MBPTrack drops significantly below this threshold. This demonstrates the advantage of the linear-time state space model over the quadratic complexity of Transformer-based architectures.  

\textbf{Computational complexity (FLOPs).}
The FLOPs of MambaTrack3D grow much more slowly with increasing input points compared to MBPTrack. While MBPTrack exhibits a sharp quadratic rise in FLOPs, MambaTrack3D maintains near-linear growth, confirming the theoretical efficiency of the proposed inter-frame propagation design.  

Importantly, the superior runtime efficiency of MambaTrack3D does not come at the cost of accuracy. As shown in the previous sections, MambaTrack3D achieves consistent improvements in success and precision across KITTI-HV and nuScenes-HV benchmarks, while simultaneously reducing computational overhead.  
These results highlight that MambaTrack3D achieves a favorable accuracy–efficiency trade-off, making it more suitable for real-time deployment in resource-constrained platforms.

\subsection{Result Visualization}
\begin{figure*}[htbp]
\centering
\includegraphics[width=0.9\linewidth]{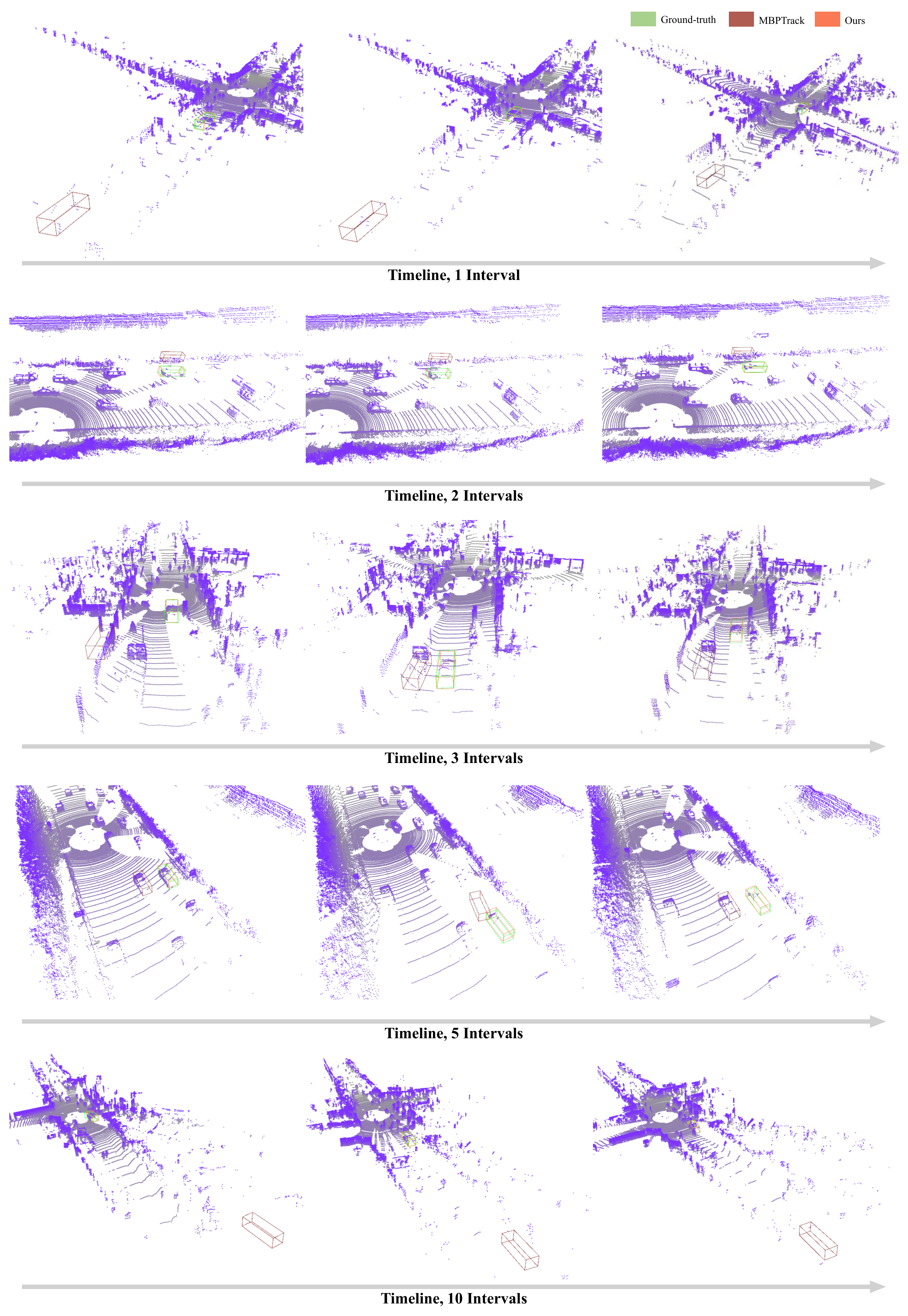}
\caption{Visualization of Tracking results on KITTI. We plot fives tracklets with different temporal interval, \ie, interval=1, 2, 3, 5, 10. MambaTrack3D is highlighted in orange, and the ground-truth in green. The best view can zoom-in. }
\label{fig:visual_car}
\end{figure*}

Fig.\ref{fig:visual_car} presents a progressive comparison between MambaTrack3D and MBPTrack across increasing temporal intervals (1, 2, 3, 5, and 10 frames). The visualization highlights how each tracker responds to growing temporal gaps and appearance variations in dynamic 3D point cloud environments.

At short intervals (1 frame), both methods closely follow the ground-truth bounding boxes, indicating reliable performance under minimal appearance change. However, as the interval increases to 2-3 frames, MBPTrack begins to accumulate positional drift, with bounding boxes gradually deviating from the true object location. In contrast, MambaTrack3D maintains accurate alignment by leveraging dynamic keypoint selection and optimal transport matching.
At 5-frame intervals, MBPTrack fails to recover from abrupt object motion (e.g., lane changes), resulting in complete tracking loss. MambaTrack3D, however, continues to track successfully by exploiting spatial distribution priors and selective historical points based on the MIP module.
Under the most challenging 10-frame interval, MBPTrack suffers from compounded errors, leading to severe misalignment. MambaTrack3D demonstrates superior robustness, preserving tracking continuity through sparse yet discriminative temporal fusion.

\subsection{Results on Normal Tracking}
Table \ref{tab:KITTI} compares MambaTrack3D with state-of-the-art (SOTA) trackers on the standard KITTI dataset, where frame-by-frame annotations are available and temporal variation is relatively small. 
Overall, MambaTrack3D achieves a mean success/precision of 70.0/87.9, which is on par with the strongest normal-scenario tracker MBPTrack (70.3/87.9) and clearly outperforms HVTrack (65.5/83.1). This demonstrates that our method, although designed for high temporal variation (HTV), maintains strong competitiveness in conventional settings. 
Compared to HVTrack, MambaTrack3D achieves significant improvements across all categories (+3.2/+5.1 for cars, +4.8/+3.5 for pedestrians), proving that the proposed propagation and grouped feature enhancement modules not only benefit HTV scenarios but also strengthen discriminability in stable frame-by-frame tracking.

When compared to conventional trackers, MambaTrack3D is highly competitive with MBPTrack, the current best-performing method, essentially matching its mean performance. 
Notably, compared to MBPTrack trained on our own devices, our OTTrack slightly surpasses it in mean success/precision rates (68.5/85.8 vs 70.0/87.9). Moreover, against other strong baselines such as CXTrack (67.5/85.3) and M2Track (62.9/83.4), MambaTrack3D shows clear improvements, particularly in the Van and Cyclist categories. 

By jointly analyzing Table \ref{tab:KITTI_HV} and Table \ref{tab:KITTI}, MambaTrack3D demonstrates dual capability: it excels in high temporal variation scenarios while remaining highly competitive in conventional frame-by-frame tracking. The results validate that our temporal propagation and feature enhancement strategies generalize well, providing robustness across both specialized and normal tracking settings.

\renewcommand{\arraystretch}{1.5} %line height
\begin{table}[tbp]
  \centering
  \caption{Comparison of MambaTrack3D with the SOTA methods on each category of the KITTI dataset.}
  \resizebox{1\linewidth}{!}{
  \begin{threeparttable}
    \begin{tabular}{c|c|c|c|c|c}
        \toprule[1.5pt]
        Methods & Car   & Pedestrian & Van   & Cyclist & Mean \\
        (Frame Number) & (6424)  & (6088)  & (1248)  & (308)   & (14068) \\
        \midrule
        P2B~\cite{P2B}   & 56.2/72.8 & 28.7/49.6 & 40.8/48.4 & 32.1/44.7 & 42.4/60.0 \\
        % LTTR~\cite{}  & 65.0/77.1 & 33.2/56.8 & 35.8/45.6 & 66.2/89.9 & 48.7/65.8 \\
        MLVSNet~\cite{mlvsnet} & 56.0/74.0 & 34.1/61.1 & 52.0/61.4 & 34.3/44.5 & 45.7/66.7 \\
        BAT~\cite{BAT}   & 60.5/77.7 & 42.1/70.1 & 52.4/67.0 & 33.7/45.4 & 51.2/72.8 \\
        PTT~\cite{PTT}   & 67.8/81.8 & 44.9/72.0 & 43.6/52.5 & 37.2/47.3 & 55.1/74.2 \\
        V2B~\cite{V2B}   & 70.5/81.3 & 48.3/73.5 & 50.1/58.0 & 40.8/49.7 & 58.4/75.2 \\
        PTTR~\cite{PTTR}  & 65.2/77.4 & 50.9/81.6 & 52.5/61.8 & 65.1/90.5 & 57.9/78.1 \\
        STNet~\cite{stnet} & 72.1/84.0 & 49.9/77.2 & 58.0/70.6 & 73.5/93.7 & 61.3/80.1 \\
        TAT~\cite{TAT}  & 72.2/83.3 & 57.4/84.4 & 58.9/69.2 & 74.2/93.9 & 64.7/82.8 \\
        M2Track~\cite{MMTrack} & 65.5/80.8 & 61.5/88.2 & 53.8/70.7 & 73.2/93.5 & 62.9/83.4 \\
        CXTrack~\cite{cxtrack} & 69.1/81.6 & 67.0/91.5 & 60.0/71.8 & 74.2/94.3 & 67.5/85.3 \\
        MBPTrack~\cite{MBPTrack} & 73.4/84.8 & 68.6/93.9 & 61.3/72.7 & 76.7/94.3 & 70.3/87.9 \\
        MBPTrack*~\cite{MBPTrack} & 72.1/83.5 & 66.6/91.4 & 57.8/68.5 & 76.0/94.4 & 68.5/85.8 \\
        \midrule
        HVTrack\cite{hvtrack} & 68.2/79.2 & 64.6/90.6 & 54.8/63.8 & 72.4/93.7 & 65.5/83.1 \\
        Ours & 71.4/84.3 & 69.4/94.1 & 64.0/74.8 & 78.3/95.2 & 70.0/87.9 \\
        \bottomrule[1.5pt]
    \end{tabular}%
    \begin{tablenotes}
        \footnotesize
        \item *indicates that the results are obtained by re-training this method on our own computer. 
    \end{tablenotes}
  \end{threeparttable}
  }
  \label{tab:KITTI}%
\end{table}%

\subsection{Ablation Studies}
\textbf{Effect of memory bank size $\rho$.}
% 本文提出了一种基于状态空间模型的点云跟踪框架，其自然地将特征提取和模板嵌入两阶段流程合并为单阶段的历史信息传播过程。在该过程中，过往历史帧的数量可能会对跟踪结果产生影响。因此，我们在KITTI数据集上做了关于记忆池容量\rho的自我对比实验。如表格所示，
This paper proposes a point cloud tracking framework based on a state space model, which naturally merges the two-stage process of feature extraction and template embedding into a single-stage historical information propagation. Considering the number of past frames stored may influence the tracking performance, we conduct a self-comparison experiment on the KITTI dataset to evaluate the effect of the memory bank capacity. 

As shown in Table~\ref{tab:memory_size}, when $\rho$ increases from 1 to 3, both success rate (SR) and precision rate (PR) improve steadily for both car and pedestrian categories. For example, in the pedestrian class, SR/PR rises from 33.7/49.2 at $\rho=1$ to 39.5/60.1 at $\rho=3$. These results indicate that incorporating a moderate number of historical frames provides richer temporal cues and enhances discriminability. 
However, when the memory size is further enlarged (\(\rho=4\) or \(\rho=5\)), performance begins to decline. For example, in the car category, the success rate drops from 62.7 at \(\rho=3\) to 59.9 at \(\rho=5\). This degradation suggests that excessive historical information introduces redundancy and noise, which may dilute the relevance of features aligned with the current frame.  

\renewcommand{\arraystretch}{1.5} %line height
\begin{table}[!ht]
    \centering
    \caption{The self-comparison results about the size of memory bank. ``SR'' and ``PR'' denotes success and precision metrics, respectively.  }
    \label{tab:memory_size}
    \belowrulesep=0pt
    \aboverulesep=0pt
    \begin{tabular}{c|cccc}
    \toprule[1.5pt]
        \multirow{2}{*}{Memory Bank Size} & \multicolumn{2}{c}{Car} & \multicolumn{2}{c}{Pedestrian} \\
        \cmidrule(lr){2-3} \cmidrule(lr){4-5}
        ~& SR(\%) & PR(\%) & SR(\%) & PR(\%)\\
    \midrule
        $\rho = 1$ &  60.3 & 70.2 & 33.7 & 49.20 \\
        $\rho = 2$ &  61.1 & 72.1 & 39.2 & 58.3 \\
        $\rho = 3$ &  62.7 & 72.8 & 39.5 & 60.1 \\
        $\rho = 4$ &  61.6 & 71.5 & 38.6 & 57.6 \\
        $\rho = 5$ &  59.9 & 71.1 & 38.9 & 57.8 \\        
    \bottomrule[1.5pt]
    \end{tabular}
\end{table}

\begin{figure}
    \centering
    \includegraphics[width=1\linewidth]{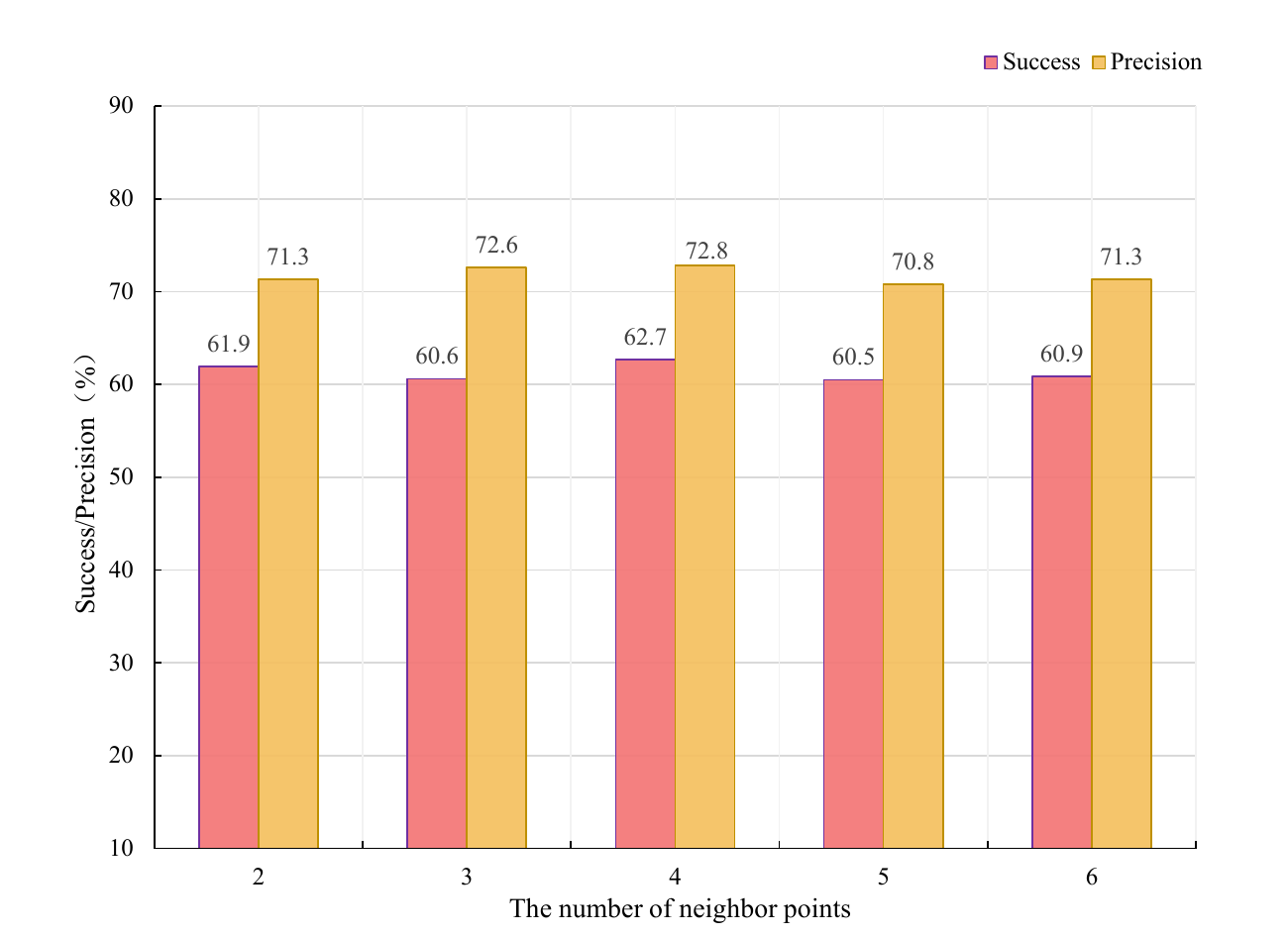}
    \caption{The impact of the number of neighbors. The results are produced on the car class of KITTI.}
    \label{fig:neighbors}
\end{figure}

\textbf{Impact of the number of neighbor point set $\mathcal{N}(q_j)$. }
% 对于每一个跟踪实例，随着时间的推移，我们按照队列数据结构来管理记忆池。然而，受视角变化和遮挡变化影响，并不是队列中的每一个历史点都能提供有利线索。为了降低记忆池里面的冗余信息，所提出的方法利用KNN在历史帧同一坐标系下来选取前K个邻居点。此处，我们考虑训练过程中邻居点数量差异对跟踪结果的影响。图给出了分别将K设置为不同值的精确率和成功率
For each tracking instance, we maintain a queue-structured memory bank over time. Due to view changes and occlusions, not every historical point is informative. To reduce redundancy, we construct a local neighbor in the same coordinate system via KNN. We investigate how the number of neighbors affects tracking performance. As shown in Figure \ref{fig:neighbors}, setting $|\mathcal{N}(q_j)|=4$ yields the highest success and precision, reaching 62.7/72.8 (Success/Precision). Setting $|\mathcal{N}(q_j)|=2$ also produces competitive results (61.9/71.3), which demonstrates that even sparse historical support is useful when points are carefully selected. The large $|\mathcal{N}(q_j)|$ expands spatial coverage of historical cues but risks incorporating points affected by pose drift or background contamination. In addition, the narrow range of success $(61 \pm 1.0)$ and precision $(\approx 71 \pm 1.0)$ across $|\mathcal{N}(q_j)|$ indicates that the pipeline is relatively robust. 

\begin{figure}
    \centering
    \includegraphics[width=1\linewidth]{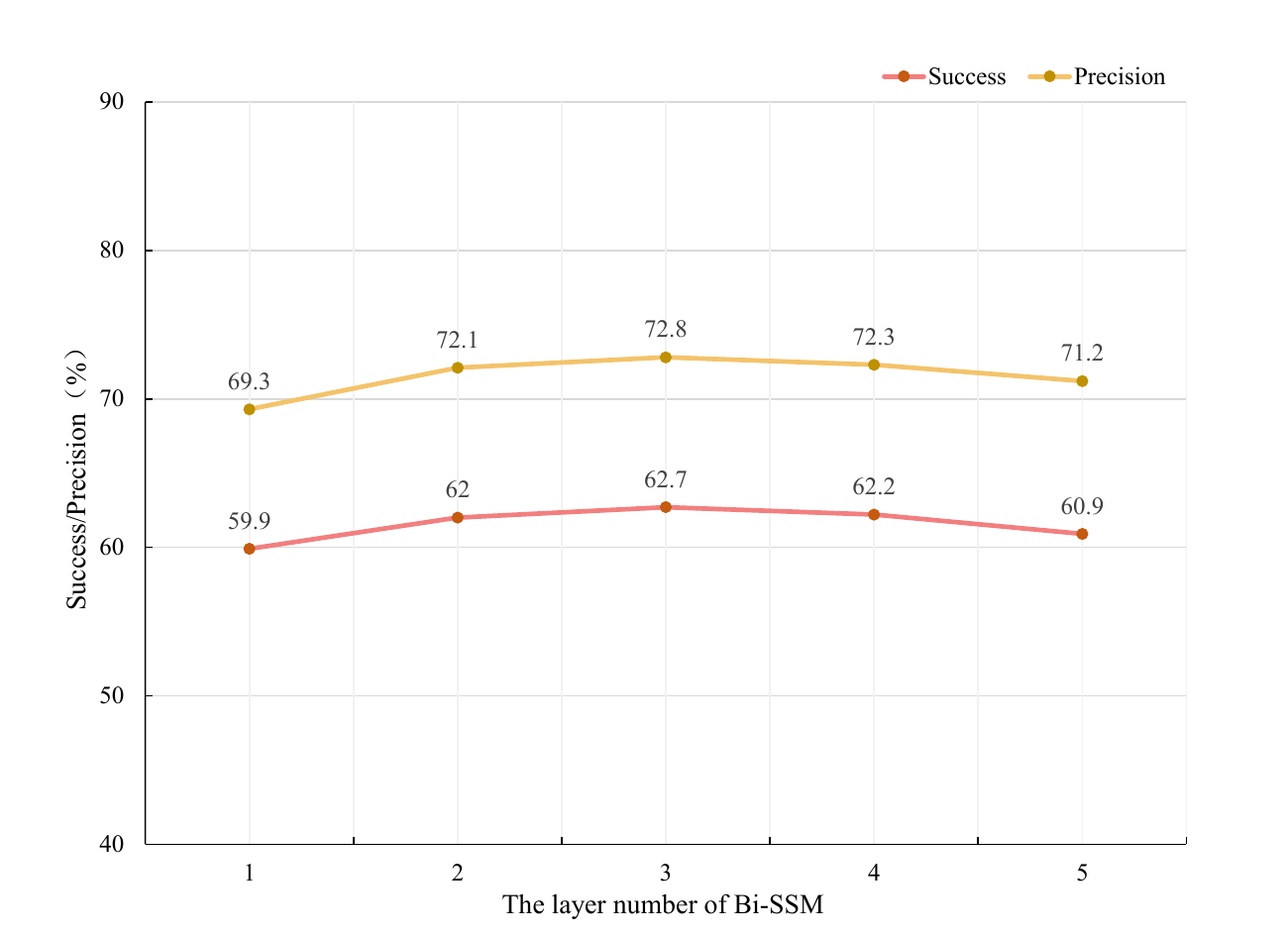}
    \caption{The self-comparison results of the layer number of Bi-SSM.}
    \label{fig:ssm_layer}
\end{figure}

\textbf{Validation of the proposed GFEM.}
% 通过在历史帧中构建局部邻域，并筛选topK个关键点可以缓解时序积累中的冗余问题。但是，由于当前帧的点云特征主要基于历史帧特征的传播方式产生，记忆池中的冗余特征会使得在邻域中计算的特征缺少判别性。鉴于此，我们提出前景和背景分组特征增强模块。此处，我们验证该模块的有效性。
To alleviate redundancy accumulated in the memory bank, we construct local neighborhoods in historical frames and select top-K keypoints. However, since the current frame features are mainly propagated from historical frames, redundant features in the memory bank may lack the discriminability of neighborhood features. To address this issue, we introduce the Grouped Feature Enhancement Module (GFEM), which explicitly separates foreground and background channels to enhance feature representation.
Table~\ref{tab:GFEM} reports the validation results of GFEM on the KITTI dataset. Incorporating GFEM consistently improves both success rate (SR) and precision rate (PR) across categories. 
For Car, SR/PR increases from 60.8/70.4 (w/o GFEM) to 62.7/72.8 (w/ GFEM). For Pedestrian, the improvement is even more pronounced, rising from 37.0/54.9 to 39.5/60.1. 

\renewcommand{\arraystretch}{1.5} %line height
\begin{table}[!ht]
    \centering
    \caption{The validation of the Grouped Feature Enhancement Module (GFEM). ``SR'' and ``PR'' denotes success and precision metrics, respectively.}
    \label{tab:GFEM}
    \belowrulesep=0pt
    \aboverulesep=0pt
    \begin{tabular}{c|cccc}
    \toprule[1.5pt]
        \multirow{2}{*}{Network Architecture} & \multicolumn{2}{c}{Car} & \multicolumn{2}{c}{Pedestrian} \\
        \cmidrule(lr){2-3} \cmidrule(lr){4-5}
        ~& SR(\%) & PR(\%) & SR(\%) & PR(\%)\\
    \midrule
        w/ GFEM  &  62.7 & 72.8 & 39.5 & 60.1 \\
        w/o GFEM &  60.8 & 70.4 & 37.0 & 54.9 \\
    \bottomrule[1.5pt]
    \end{tabular}
\end{table}

\textbf{Impact of the layer number of Bi-SSM.}
% 本文采用了双向状态空间模型在不同帧间点云进行特征传播。此处，我们考察了网络结构中不同层数对最终跟踪性能的影响。
We employ a bidirectional state space model (Bi-SSM) to propagate features across frames. To assess architectural depth, we vary the number of Bi-SSM layers from 1 to 5 and report success and precision. As shown in Fig. \ref{fig:ssm_layer}, three Bi-SSM layers provide the best balance between temporal information propagation and robustness. A single layer underfits temporal relations (59.9/69.3), while 2 layers substantially improve both metrics (62.0/72.1), showing clear gains from modest depth. Increasing to 4 layers maintains precision (72.3) but slightly reduces success (62.2), and 5 layers further degrades both metrics (60.9/71.2). This suggests deeper stacks introduce redundant or misaligned temporal cues that weaken correspondence quality. We adopted $L=3$ as the default configuration.

\renewcommand{\arraystretch}{1.5} %line height
\begin{table}[!ht]
    \centering
    \caption{The self-comparison results of only using geometric features and semantic mask features. ``SR'' and ``PR'' denotes success and precision metrics, respectively.}
    \label{tab:feat}
    \belowrulesep=0pt
    \aboverulesep=0pt
    \begin{tabular}{cc|cccc}
    \toprule[1.5pt]
        \multirow{2}{*}{Geometric Feat.} & \multirow{2}{*}{Mask Feat.} & \multicolumn{2}{c}{Car} & \multicolumn{2}{c}{Pedestrian} \\
        \cmidrule(lr){3-4} \cmidrule(lr){5-6}
        ~&~& SR(\%) & PR(\%) & SR(\%) & PR(\%)\\
    \midrule
        \checkmark  & ~          &  61.9 & 71.8 & 37.0 & 55.2 \\
        ~           & \checkmark &  52.0 & 62.1 & 36.6 & 56.5 \\
        \checkmark  & \checkmark &  62.7 & 72.8 & 39.5 & 60.1 \\
    \bottomrule[1.5pt]
    \end{tabular}
\end{table}
\textbf{Effect of geometric and mask semantic information.} 
% 在公式6中，我们同时考虑了点云的几何特征以及掩码的语义特征，为了说明二者的相辅相成。我们研究了只使用单一特征类型对模型训练结果的影响。
In Eq. (6), both geometric features of point clouds and semantic mask features are fused to enhance discriminability. To validate their complementary roles, we conduct an ablation study by training the model with only one type of feature at a time.
When geometric and semantic features are integrated, the model achieves the best performance (62.7/72.8 on Car and 39.5/60.1 on Pedestrian). But when solely relying on geometric information, the model achieves 61.9/71.8 (SR/PR) on Car and 37.0/55.2 on Pedestrian. This indicates that geometric cues provide a solid foundation for rigid objects but are less effective for small or deformable targets.
With mask features alone, performance drops significantly for Car (52.0/62.1) but remains comparable for Pedestrian (36.6/56.5). This suggests that semantic cues are helpful for distinguishing small targets, especially in cluttered pedestrian scenes, but insufficient to capture precise geometric structures.

\section{Conclusion}
In this work, we presented MambaTrack3D, a LiDAR-based tracking framework tailored for high temporal variation scenarios. By leveraging the state space model Mamba, our inter-frame propagation module achieves linear-time complexity and effectively exploits spatial–temporal relations across historical frames. To further reduce redundancy, we introduced a grouped feature enhancement module, which enhances discriminability by explicitly modeling foreground–background semantics.
Comprehensive experiments on KITTI-HTV and nuScenes-HTV benchmarks validate the effectiveness of our approach. MambaTrack3D consistently surpasses the HTV baseline HVTrack  under challenging temporal gaps, while maintaining competitive performance in the normal tracking setting.  These results highlight its dual capability: excelling at temporal variation and remaining efficient computation.

Future work will explore extending our framework to class-agnostic \cite{ClassAgnostic_Tracking} or class-unified tracking \cite{TrackAny3D}, and integrate adaptive memory management strategies, further enhancing robustness in real-world autonomous driving and robotic perception systems.

\section*{Acknowledgment}
This work is supported in part by the National Natural Science Foundation of China (Grant No. 62301562), and the China Postdoctoral Science Foundation (Grant No. 2023M733756).

\ifCLASSOPTIONcaptionsoff
  \newpage
\fi



\bibliographystyle{IEEEtran}
\bibliography{mybib}

\vfill


\end{document}